\documentclass{article}
\usepackage{algorithm}
\usepackage{algpseudocode}
\usepackage{amsmath}
\usepackage{graphicx}
\usepackage{array}
\usepackage{subcaption}
\usepackage[numbers,sort&compress]{natbib}
\usepackage{authblk}
\usepackage{hyperref}

\usepackage[preprint]{neurips_2024}

\usepackage[utf8]{inputenc} 
\usepackage[T1]{fontenc}    
\usepackage{hyperref}       
\usepackage{url}            
\usepackage{booktabs}       
\usepackage{amsfonts}       
\usepackage{nicefrac}       
\usepackage{microtype}      
\usepackage{xcolor}         

\hypersetup{
    colorlinks=true,    
    linkcolor=blue,    
    urlcolor=blue,     
    citecolor=blue     
}

\title{DiffFluid: Plain Diffusion Models are Effective Predictors of Flow Dynamics}

\author[1,2]{\textbf{Dongyu Luo}}
\author[1,5]{\textbf{Jianyu Wu}}
\author[3]{\textbf{Jing Wang}}
\author[3]{\textbf{Hairun Xie}}
\author[4]{\textbf{Xiangyu Yue}}
\author[1,4]{\textbf{Shixiang Tang}}

\affil[1]{Shanghai Artificial Intelligence Laboratory}
\affil[2]{The University of Hong Kong}
\affil[3]{COMAC Shanghai Aircraft Design \& Research Institute}
\affil[4]{The Chinese University of Hong Kong}
\affil[5]{Beijing University of Aeronautics and Astronautics}

\begin{document}

\maketitle

\begin{abstract}
We showcase the plain diffusion models with Transformers are effective predictors of fluid dynamics under various working conditions, \emph{e.g.,} Darcy flow and high Reynolds number. Unlike traditional fluid dynamical solvers that depend on complex architectures to extract intricate correlations and learn underlying physical states, our approach formulates the prediction of flow dynamics as the image translation problem and accordingly leverage the plain diffusion model to tackle the problem. This reduction in model design complexity does not compromise its ability to capture complex physical states and geometric features of fluid dynamical equations, leading to high-precision solutions. In preliminary tests on various fluid-related benchmarks, our DiffFluid achieves consistent state-of-the-art performance, particularly in solving the Navier-Stokes equations in fluid dynamics, with a relative precision improvement of \textbf{+44.8\%}. In addition, we achieved relative improvements of \textbf{+14.0\%} and \textbf{+11.3\%} in the Darcy flow equation and the airfoil problem with Euler’s equation, respectively. Code will be released at \url{https://github.com/DongyuLUO/DiffFluid} upon acceptance.
\end{abstract}

\section{Introduction}

Computational Fluid Dynamics (CFD) has gathered increasing attention in both academic and engineering researches, which aims at developing models to simulate fluid flows to maximize the throughput in chemical plants, optimize the energy yield of wind turbines, or improve the efficiency of aircraft engines. The widespread use of these simulations make their acceleration through machine learning highly influential. 

The traditional CFD simulations~\cite{ames2014numerical, mazumder2015numerical} consist of a discretized geometry represented as a grid or mesh, boundary conditions that specify the position of walls and inlets, and initial conditions that provide a known state of the flow, and predict the flow of fluids using numerical solvers that solve partial differential equations, such as Naiver-Stokes equations:

\begin{equation} \label{eq:NS_eq}
    \frac{\partial \mathbf{u}}{\partial t} = \nu \nabla^2 \mathbf{u} - \frac{1}{\rho} \nabla p -(\mathbf{u}\cdot \nabla) \mathbf{u},
\end{equation}
forward over a given time interval. The partial differential equation above describes the relationship between velocity $\mathbf{u}$ and pressure $p$ for a viscous fluid with kinematic viscosity $\nu$ and constant density $\rho$. For liquids, Eq.~\ref{eq:NS_eq} is further constrained by the incompressibility assumption $\nabla \cdot \mathbf{u} = 0$. Despite their high precision, large-scale simulations can be very slow. For example, it takes 30 minutes to simulate flow maps with 12,600 cells \cite{chen2012optimization}.


With the rise of deep learning, a prominent pathway to accelerate the fluid flow simulations is to emulate numerical solvers using deep learning models~\cite{li2020fourier, li2020neural, kovachki2023neural, lu2021learning, raissi2019physics}. Typical methods follow a paradigm of deterministic models, predicting the flow map after a given time interval based on the initial flow map. Despite its success, this paradigm still suffers from the following drawbacks. First, current deterministic models rely on U-Net architectures supervised by the target flow maps, which overlook the multiscale nature of flow dynamics. Second, these models often fail to accurately resolve nonlinear interactions and complex boundary conditions, as well as sharp interfaces within the flow field, all of which are critical for achieving reliable predictions in fluid dynamics. These limitations highlight the need to shift toward generative models in flow prediction.

In this paper, we demonstrate that plain diffusion models can be repurposed as efficient and powerful flow predictors of fluids. Our intuition behind is that flow prediction can be reformulated as an image-to-image translation problem, a task at which generative models have proven to excel. The key is to model multiscale dynamics within a single generative model. Specifically, we propose a fluid flow predictor based on plain diffusion models, called DiffFluid. This predictor combines diffusion models with Transformer architecture to create a joint probability distribution between input conditions and output solutions, effectively capturing interdependencies in discrete spaces. During inference phase, we apply standard diffusion models. Additionally, we have designed two strategies to significantly enhance solution accuracy: the multi-resolution noise strategy and the multi-loss strategy. This design not only maintains structural simplicity but also significantly improves solution accuracy compared to previous state-of-the-art solvers. In summary, our contributions can be summarized as follows:

\begin{figure}[t]
    \centering
    \includegraphics[width=1\textwidth]{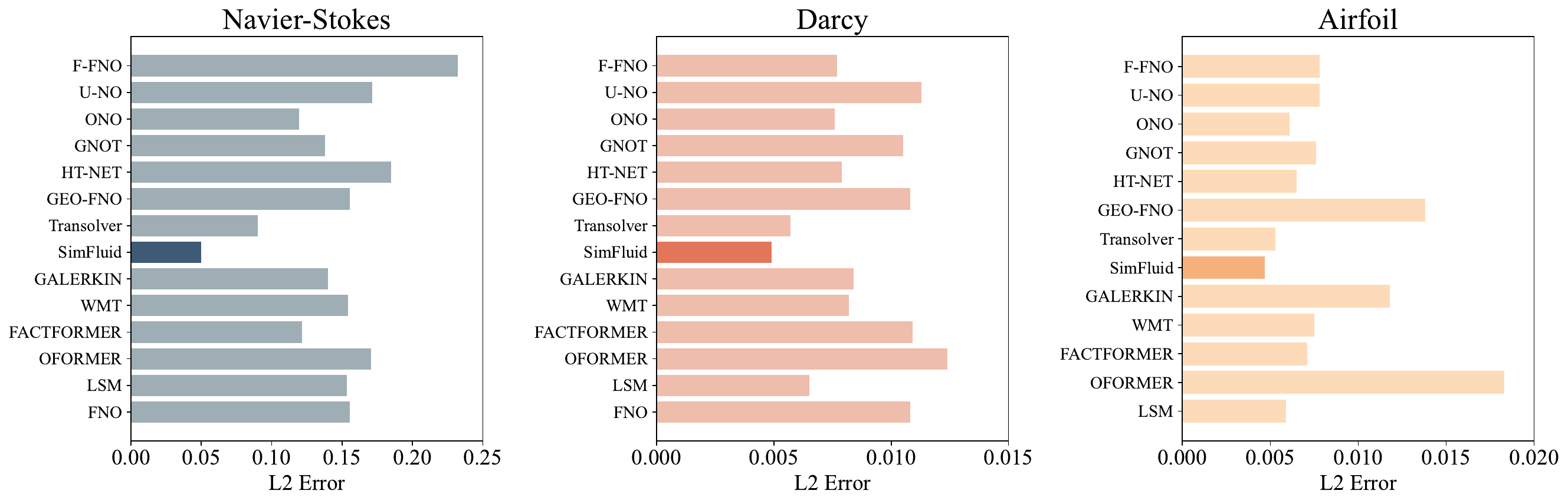} 
    \caption{From left to right, the $L2$ error of different models on three benchmarks: Navier-Stokes equation, Darcy flow equation and airfoil problem with Euler's equation.}
    \label{fig:pre}
\end{figure}


\begin{itemize}
    \item We demonstrate that plain diffusion models are effective fluid flow predictors, which significantly simplifies the previously complex solvers for fluid dynamical equations. 
    
    \item We propose a multi-resolution noise strategy and a multi-loss strategy that effectively capture multi-resolution dynamics and sharp geometric boundaries in the solution domain of fluid dynamical equation, which are crucial for model accuracy, thereby significantly enhancing the solution precision.
\end{itemize}

Our model improves by \textbf{+44.8\%} over previous state-of-the-art methods~\cite{wu2024transolver, wu2022flowformer, li2022transformer, hao2023gnot, xiao2023improved} in solving the Navier-Stokes equations for fluid dynamics. It also achieves performance gains in the Darcy flow case, where we tested different resolutions, demonstrating both its accuracy and versatility. Specifically, DiffFluid consistently achieves state-of-the-art performance in the Darcy flow case at resolutions ranging from $41 \times 41$ to $141 \times 141$, with an average improvement of approximately \textbf{+14.4\%} over the second-best model.

\section{Related Work}

\subsection{Diffusion model}
Diffusion models are widely used in various tasks  including image generation \cite{ho2022cascaded}, restoration \cite{xia2023diffir}, and super-resolution \cite{li2022srdiff}, as well as text-to-image \cite{ruiz2023dreambooth}, video \cite{ho2022video}, and audio generation \cite{liu2023audioldm}, among others. Additionally, diffusion models can be used for data augmentation \cite{trabucco2023effective} to enhance the robustness of machine learning models. Due to their flexibility and high-quality generation capabilities, diffusion models are widely applied across multiple fields. The Denoising Diffusion Probabilistic Model (DDPM) \cite{ho2020denoising} is a widely used diffusion model. It gradually adds Gaussian noise to data, turning it into pure noise. Then, the model is trained to learn how to progressively remove the noise to recover high-quality samples. This process includes diffusion and reverse diffusion. The model generates new samples by predicting noise in the denoising steps, ultimately achieving effectively achieving tasks like image generation. Furthermore, diffusion models have been used to generate fluid-related datasets, as demonstrated in \cite{lienen2023zero}. This further demonstrates the potential of diffusion models in solving fluid dynamic. Building on this, we propose a solver based on DDPM to tackle fluid dynamics equations.

\subsection{Deep learning fluid dynamic solver}
For a long time, various numerical methods have been widely used to solve fluid dynamical equations, including the Finite Difference Method \cite{smith1985numerical}, Finite Element Method \cite{johnson2009numerical}, Finite Volume Method \cite{moukalled2016finite}, and Spectral Methods \cite{gottlieb1977numerical}. With the advent of deep learning, two types of deep learning paradigms for solving fluid dynamical equation have emerged. One class is learning-based PINNs represented by \cite{raissi2019physics}, and the other is data-driven based neural operators represented by \cite{li2020fourier}.

\textbf{Physics-informed neural networks}\quad This paradigm was proposed by \cite{raissi2019physics}. It \cite{yang2021b, ren2022phycrnet, yu2022gradient} uses fluid dynamical equation constraints, such as equations, boundary conditions, and initial conditions, as a loss function. A self-supervised learning approach trains the neural network, allowing the model's output to gradually satisfy the fluid dynamical equation constraints and ultimately achieve an approximate solution. However, this method often relies heavily on network optimization, which limits its scalability. 

\textbf{Neural operators}\quad Another paradigm establishes the mapping between input and output in fluid dynamical equations solving tasks using neural operators. For example, in the task of solving the Navier-Stokes equations, the current state of the flow field is used as input to predict the future state of the flow field \cite{li2020fourier}. The foundation of this paradigm is usually credited to the Fourier Neural Operator (FNO) proposed by \cite{li2020fourier}. The main concept of this operator is to approximate integration using linear projections in the Fourier domain. Building on this foundation, various improvements have emerged. For example, U-FNO \cite{wen2022u} and U-NO \cite{rahman2022u} proposed using the U-Net \cite{ronneberger2015u} architecture to enhance the performance of the FNO. WMT \cite{gupta2021multiwavelet} introduced multiscale wavelet bases to capture the complex relationships between different scales. To improve the efficiency of the model, F-FNO \cite{tran2021factorized} utilizes factorization in the Fourier domain. In addition to address the high dimensional complexity problem present in fluid dynamical, LSM\cite{wu2022flowformer} uses spectral methods \cite{gottlieb1977numerical} in the learned latent space.

In particular, due to the boom of Transformer \cite{vaswani2017attention}, it is also utilized in the task of solving fluid dynamical equations. HT-Net \cite{liu2022ht} improves the performance of the model in capturing multiscale spatial correlations by incorporating Swin Transformer \cite{liu2021swin} and multigrid methods \cite{wesseling1995introduction}. OFormer \cite{li2022transformer}, GNOT \cite{hao2023gnot}, and ONO \cite{xiao2023improved} utilize current advanced Transformer architectures, such as the Reformer \cite{kitaev2020reformer}, Performer \cite{choromanski2020rethinking}, and Galerkin Transformer \cite{cao2021choose}, applying attention between grid points. Recently, Transolver \cite{wu2024transolver} proposes to construct mappings of inputs to outputs by learning the intrinsic physical state of the fluid dynamical equation captured by learnable slices. Nevertheless, the previous methods often lead to complex model structures to capture the geometric and physical states of fluid dynamics, limiting potential improvements. In contrast, DiffFluid achieves state-of-the-art performance with a simple architecture and no fine-tuning, effectively capturing complex relationships in fluid dynamics. With further optimization, its performance is expected to improve significantly and could be extended to a wider range of fluid dynamical equations, with the potential to become a high-precision general solver.

\section{Method}
\subsection{Diffusion generative formulation}
We approach solving fluid dynamical equations as a conditional denoising diffusion generation task. We train DiffFluid to model the conditional distributions \( D(y|x) \), where \( y \in \mathbb{R}^{C_o} \) represents the output of the fluid dynamical equation and \( C_o \) denotes the dimension of the output space. The condition \( x \in \mathbb{R}^{C_i} \) represents the input of the fluid dynamical equation, with \( C_i \) indicating the dimension of the input space.

\begin{figure}[t]
    \centering
    \includegraphics[width=1\textwidth]{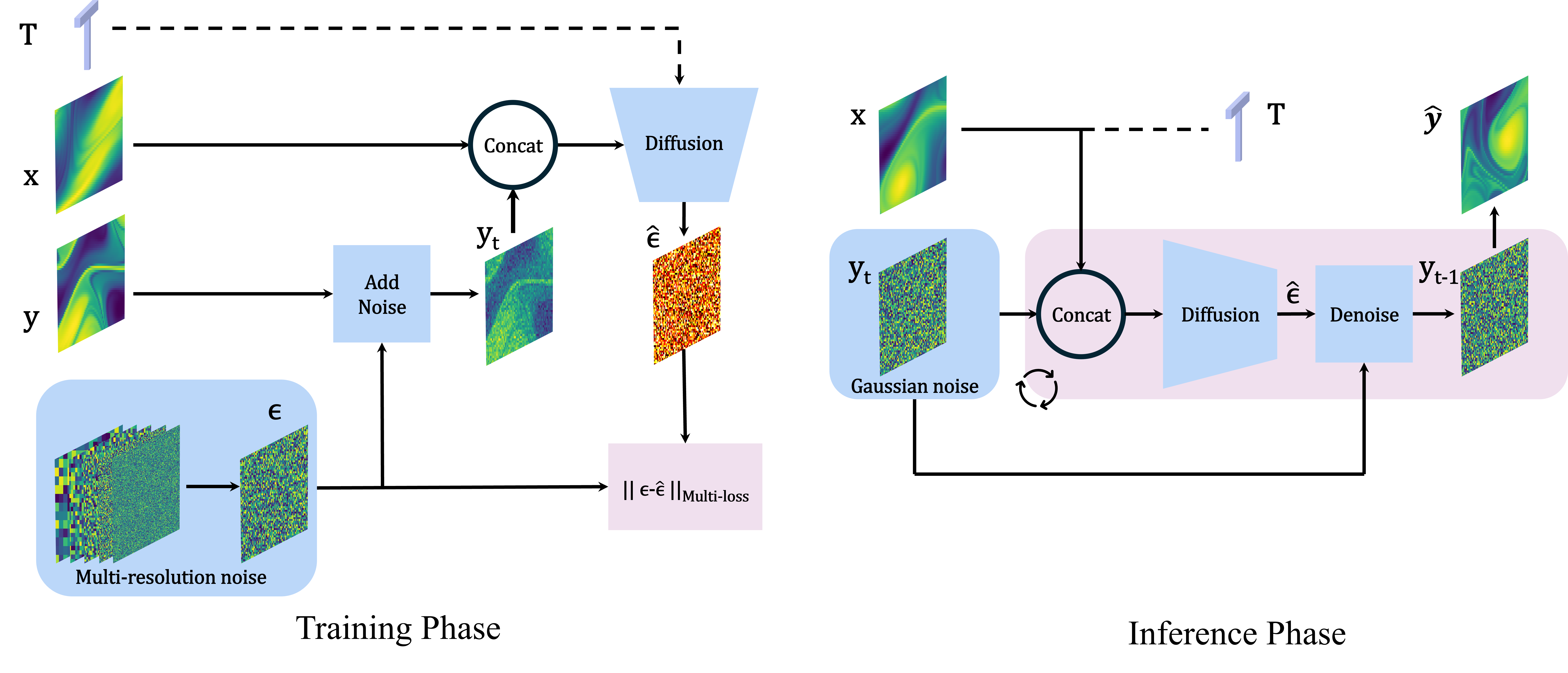}
    \caption{Left: Structure diagram of the DiffFluid training phase. Right: Structure diagram of the DiffFluid inference phase.}
    \label{fig:train_infer}
\end{figure}

During forward diffusion, starting from the conditional distribution at $y_0 := y$,  Gaussian noise is gradually added over time steps $t\in \left \{ 1,2,3, ...,T\right \}$ to obtain the noisy samples $y_t$ as
\begin{equation}
    y_t = \sqrt{\bar{\alpha_t}}y_0 + \sqrt{1-\bar{\alpha_t}}\epsilon 
\end{equation}
where $\epsilon\sim\mathcal{N}\left ( 0,\mathit{I}\right )$, $\bar{\alpha_t} := \textstyle\prod_{s=1}^{t}1-\beta_s$, and $\left \{ \beta_1,\beta_2,\beta_3, ...,\beta_T\right \}$ represents the variance schedule of a process over T steps. In the reverse process, the conditional denoising model $\epsilon_\theta\left(\cdot\right)$, which is parameterized by learned parameters $\theta$, progressively removes noise from $y_t$ to obtain $y_{t-1}$.

During training, parameters $\theta$ are updated by taking a data pair $\left(x, y\right)$ from the training data. At a random time step $t$, noise $\epsilon$ is applied to $y$, and the noise estimate $\hat{\epsilon} = \epsilon_\theta\left(y_t, x, t\right)$ is calculated. One of the denoising objective function is minimized, with a typical standard noise objective $\mathcal{L}$ as follows:
\begin{equation}
    \mathcal{L} = \mathbb{E}_{y_0, \epsilon \sim\mathcal{N}\left(0,\mathit{I}\right), t \sim \mathit{U}\left(\mathit{T}\right)}\|\epsilon-\hat{\epsilon}\|^2
\end{equation}

During inference, $y:=y_0$ is reconstructed from a normally distributed variable $d_T$ by applying the learned denoiser $\epsilon_\theta\left(y_t, x, t\right)$ iteratively.

\subsection{Network architecture}

\textbf{Problem setup}\quad We defined the fluid dynamical equations over an input domain $\Omega \subset \mathbb{R}^{C_{x_g}}$, where $C_{x_g}$ represents the dimension of the input space. And we usually discretize $\Omega$ into $N$ points $x_g \in \mathbb{R}^{N \times C_{x_g}}$. The objective is to estimate the output of fluid dynamical equations the  based on the input geometries $x_g$ and the physical quantities $x_q \in \mathbb{R}^{N \times C_{x_q}}$ observed at these points. Note that $x_q$ is optional for some fluid dynamical equations task. In these cases, $x_g$ and $x_q$ together form the input $x \in \mathbb{R}^{N \times C_x}$ for the fluid dynamical equation, where $C_x = C_{x_g}+C_{x_q}$. And we hope to predict the output $y \in \mathbb{R}^{N\times C_y}$ of the fluid dynamical equation by inputting $x$ or $x_g$ into the model. In addition, we need to consider that for non-stationary fluid dynamical equations, we must introduce an additional time variable $T$, whereas stationary fluid dynamical equations do not require this consideration \cite{evans2022partial}.

\textbf{Training phase}\quad The overview of the inference pipeline is presented in the Figure \ref{fig:train_infer}. In order to predict the output $y$ from the input $x$ of fluid dynamical equation, our model learns the joint distribution of the fluid dynamical equation inputs and outputs. We propose an efficient training phase. Unlike traditional image generation tasks, solving fluid dynamical equations requires high precision, and the input size for fluid dynamical equations is often smaller compared to image generation. Therefore, we abandon the step of transforming fluid dynamical equations into latent space before performing the diffusion process \cite{rombach2022high}. First, we randomly select the input $x$ and its corresponding output $y$ from the training set of the fluid dynamical equations, and then add multi-resolution noise \ref{sec:design}  to $y$. Next, we concantenate the noisy $y_t \in \mathbb{R}^{N\times C_y} $ and $x$ along the feature dimension to obtain $s \in \mathbb{R}^{N\times C_s}$, where $C_s = C_x + C_y$. Finally, we input $s$ into the diffusion transformer to predict the noise $\hat{\epsilon}$, using the difference between $\hat{\epsilon}$ and $\epsilon$ to guide the training process.

\textbf{Diffusion transformer}\quad When inputting $s$ into the diffusion with transformer, the first step is to perform patch embedding on $s$ At the same time, for non-stationary fluid dynamical equations, it is necessary to embed the time $T \in \mathbb{R}^{N \times C_s}$ from the fluid dynamical equation and the time step $t \in \mathbb{R}^{N \times C_s}$ used for the diffusion process. Both embeddings are the combined through linear addition to form $T' \in \mathbb{R}^{N \times C_s}$. For stationary fluid dynamical equations, only the time step $t$ needs to be considered. After that, the embedded representations $s'\in \mathbb{R}^{N \times C_s}$ and $T'\in \mathbb{R}^{N \times C_s}$ are input into the standard DiT Block with adaLN-Zero \cite{peebles2023scalable} for training. Finally, the predicted noise $\hat{\epsilon}$ is derived from the operations of layer normalization followed by linear transformation and reshaping.

\textbf{Inference phase}\quad The overview of the inference pipeline is presented in Figure \ref{fig:train_infer}. In the inference process of DiffFluid, it start with sampling from a standard Gaussian distribution. Although using multi-resolution noise \ref{sec:design}  was found to yield better results during training, employing a sample Gaussian distribution during inference significantly reduces the randomness of the generated results. This enhances the efficiency and consistency of the inference, ensuring that the generated samples are more reliable and stable. Next, the Gaussian noise is combined with the input conditions of the fluid dynamical equation and fed into the diffusion transformer. After executing the schedule of time steps, we gradually denoise to ultimately generate the solution corresponding to the fluid dynamical equation. It is important to note that for non-stationary fluid dynamical equations, the time variable $T$ need to be considered.

\subsection{Detailed optimization design} \label{sec:design}
\textbf{Multi-resolution noise}\quad During the training process, we found that many solutions of fluid dynamical equations exhibit multiple sharp feature surfaces. Existing fluid dynamical equation solvers have struggled to effectively address this issue. Moreover, the characteristics of diffusion models tend to produce overly smooth fluid dynamical equation solutions, neglecting these important sharp features.

To address this, we propose using multi-resolution noise \cite{whitaker2023multi} to replace the Gaussian noise in standard DDPMs. This multi-resolution noise is constructed by overlaying multiple scales of random Gaussian noise, which is then adjusted to the resolution required by the diffusion transformer model. By appropriately balancing the low-frequency and high-frequency components of the noise, we achieve high fidelity of sharp feature surfaces while maintaining overall structural integrity and accuracy.

Additionally, we incorporate an annealing operation to refine the noise application. This annealing process helps to gradually reduce the noise strength over time, enhancing the clarity of the sharp features. Together, the combination of multi-resolution noise and the annealing operation not only addresses the challenge of capturing sharp features but also accelerates convergence speed during training, leading to more efficient and effective fluid dynamical equation solving.

\begin{algorithm}
    \caption{Multi-Resolution Noise with Annealing for fluid dynamical equation Solving}
    \begin{algorithmic}[1]
        \State \textbf{Input:} Number of scales $K$, weights $\alpha_i$, standard deviations $\sigma_i$, upsampling factor $r$, timesteps $T$
        \State \textbf{Output:} Multi-resolution Noise $n_{\text{input}}$ for diffusion model

        \Function{GenerateMultiResolutionNoise}{$x$, $t$}
            \State $N_{\text{multi}}(x) \gets 0$ \Comment{Initialize multi-resolution noise}
            \For{$i = 1$ \textbf{to} $K$}
                \State $N_i(x) \sim N(0, \sigma_i^2)$ \Comment{Generate Gaussian noise for scale $i$}
                \State $N_{\text{multi}}(x) \gets N_{\text{multi}}(x) + \alpha_i N_i(x)$ \Comment{Combine weighted noise}
            \EndFor
            \State \Return $N_{\text{multi}}(x)$
        \EndFunction

        \Function{AdjustResolution}{$N_{\text{multi}}, r$}
            \State $U(x) \gets \text{UpsampleOperation}(N_{\text{multi}}, r)$ \Comment{Adjust to required resolution}
            \State \Return $U(x)$
        \EndFunction

        \Function{AnnealingNoise}{$N_{\text{multi}}, t, T$}
            \State $strength \gets 1.0 - \left(\frac{t}{T}\right)$ \Comment{Calculate noise strength}
            \State $N_{\text{annealed}}(x) \gets strength \cdot N_{\text{multi}}(x)$ \Comment{Apply annealing}
            \State \Return $N_{\text{annealed}}(x)$
        \EndFunction

        \State $N_{\text{multi}} \gets \text{GenerateMultiResolutionNoise}(x, t)$ \Comment{Generate multi-resolution noise}
        \State $N_{\text{annealed}} \gets \text{AnnealingNoise}(N_{\text{multi}}, t, T)$ \Comment{Apply annealing to noise}
        \State $n_{\text{input}} \gets \text{AdjustResolution}(N_{\text{annealed}}, r)$ \Comment{Adjust noise to model's resolution}

        \State \Return $n_{\text{input}}$ \Comment{Final multi-resolution noise for diffusion model}
    \end{algorithmic}
\end{algorithm}

\textbf{Multi-loss strategy} \quad Given the importance of precise modeling in generating solutions to  fluid dynamical equations, accurately capturing multiple sharp features is crucial for the overall accuracy of the model. Traditional diffusion models typically use mean squared error (MSE) as the loss function, aiming to minimize the subtle differences between the generated solution and the true solution. However, MSE is highly sensitive to large errors, such as sharp discontinuities at boundaries, which leads the model to produce smoother boundaries and hinders the accurate capture of multiple sharp features in the fluid dynamical equation solution.

To address this issue, we propose a multi-loss strategy that builds on MSE by additionally incorporating absolute error loss (L1 loss). This strategy offers the following advantages:

\begin{itemize}
    \item \textbf{Robustness to Outliers}: L1 loss is less sensitive to large errors than MSE, meaning that when handling the boundaries of the solution, L1 loss will not overly smooth them, thereby preserving more details.
    
    \item \textbf{Boundary Sharpness}: L1 loss promotes larger numerical differences in model outputs, resulting in clearer and sharper boundaries. In contrast, MSE focuses on minimizing the sum of squared errors, which tends to produce smoother boundaries.
    
    \item \textbf{Sparse Feature Selection}: L1 loss can promote the sparsity of model parameters, highlighting important features and enhancing the overall clarity of the solution.
\end{itemize}

By effectively combining MSE and L1 loss through this weighted strategy, we can more efficiently generate multiple sharp features in fluid dynamical equation solutions, thereby significantly improving the accuracy of the generated solutions.

\section{Experiment}

We conduct experiments to validate DiffFluid using fluid dynamical equations, focusing on the Navier-Stokes equations, Darcy flow equations and an industrial airfoil problem with Euler's equations. All tests are performed on a single NVIDIA A100 40G GPU.

\subsection{Fluid dynamical equation problem settings}
\textbf{Navier-Stokes equation}\quad  In this paper, we consider the incompressive and viscous 2-d Navier-Stokes equation in vorticity form on the unit torus.
\begin{equation}
\begin{aligned}
    \partial_t w(x, t) + u(x, t) \cdot \nabla w(x, t) &= \nu \Delta w(x, t) + f(x), \quad &x &\in (0, 1)^2, \; t \in (0, T] \\
    \nabla \cdot u(x, t) &= 0, \quad &x &\in (0, 1)^2, \; t \in [0, T] \\
    w(x, 0) &= w_0(x), \quad &x &\in (0, 1)^2 
\end{aligned}
\end{equation}
Here $w = \Delta \times v$ is the vorticity, $v \left(x,t\right)$ is the velocity at $x$ at time $t$, and $f \left(x\right)$ is the forcing function. Solving the Navier-Stokes equations at high Reynolds numbers has always been a challenge. Therefore, we set the viscosity to $\nu = 1 \times 10^{-5}$, corresponding to a Reynolds number of $Re = \frac{1}{\nu} = 10^5$. In studying the Navier-Stokes equations, we are typically interested in predicting future states from the current state. Thus, this experiment is set to predict the future state $w_T$ from the current state $w_0$, with time $T$ set to $10$.

\begin{figure}[t]
    \centering
    \includegraphics[width=0.7\textwidth]{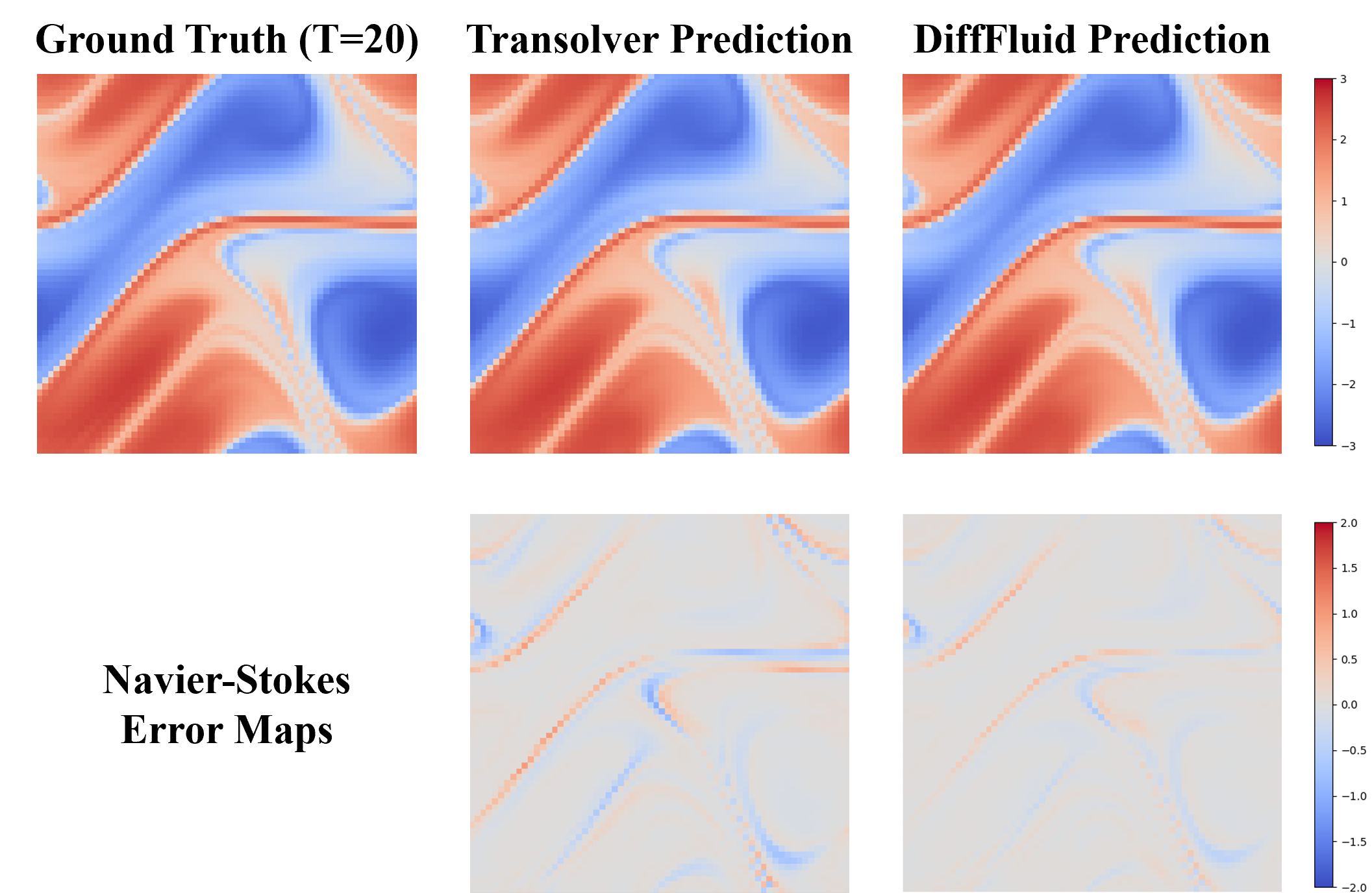} 
    \vspace{1em}
    \caption{A comparison of DiffFluid with the previous best model, Transolver, on the Navier-Stokes equation benchmark. Both prediction results and error maps are provided.}
    \label{fig:ns}
\end{figure}

\textbf{Darcy flow equation} \quad  In this paper, we validate the DiffFluid under the steady-state of the 2D Darcy flow equation on the unit box.
\begin{equation}
\begin{aligned}
    -\Delta \cdot (a(x)\Delta u(x)) &= f(x), \quad &x &\in (0, 1)^2 \\
    u(x) &= 0, \quad &x &\in \partial(0, 1)^2
\end{aligned}
\end{equation}
Here $a$ is the diffusion coefficient and $f$ is the forcing function. In this experiment, we aim to solve for $u$ using the input $a$.

\textbf{Airfoil problem with Euler’s equation}\quad In this problem we consider subsonic flow over an aerodynamic wing with governing Euler equations, as follows.

\begin{equation}
\begin{aligned}
    \frac{\partial \rho^f}{\partial t} + \nabla \cdot (\rho^f \mathbf{v}) &= 0, \\
    \frac{\partial \rho^f \mathbf{v}}{\partial t} + \nabla \cdot \left( \rho^f \mathbf{v} \otimes \mathbf{v} + p \mathbb{I} \right) &= 0, \\
    \frac{\partial E}{\partial t} + \nabla \cdot \left( (E + p) \mathbf{v} \right) &= 0,
\end{aligned}
\end{equation}

where $\rho^f$ is the  fluid density, $\mathbf{v}$ is the velocity vector, $p$ is the pressure, and $E$ is the total energy. And the viscous is ignored. We set the far-field boundary condition is $\rho_\infty = 1.0$,\quad$p_\infty = 1.0$,\quad$M_\infty = 0.8$,\quad $\alpha = 0$ where $M_\infty$ is the Mach number and $\alpha$ is the angle of attack, and at the airfoil, no-peneration condition is imposed.

\begin{table}[h]
\centering
\resizebox{1\textwidth}{!}{
\begin{tabular}{@{}lccclccc@{}}
\toprule
Model & Navier-Stokes & Darcy & Airfoil & Model & Navier-Stokes & Darcy & Airfoil \\
\midrule
FNO \cite{li2020fourier} & 0.1556 & 0.0108 & / & HT-NET \cite{liu2022ht} & 0.1847 & 0.0079 & 0.0065 \\
LSM \cite{wu2022flowformer} & 0.1535 & 0.0065 & 0.0059 & GNOT \cite{hao2023gnot} & 0.1380 & 0.0105 & 0.0076 \\
OFORMER \cite{li2022transformer} & 0.1705 & 0.0124 & 0.0183 & ONO \cite{xiao2023improved} & 0.1195 & 0.0076 & 0.0061 \\
FACTFORMER \cite{li2024scalable} & 0.1214 & 0.0109 & 0.0071 & U-NO \cite{rahman2022u} & 0.1713 & 0.0113 & 0.0078 \\
WMT \cite{gupta2021multiwavelet} & 0.1541 & 0.0082 & 0.0075 & F-FNO \cite{tran2021factorized} & 0.2322 & 0.0077 & 0.0078 \\
GALERKIN \cite{cao2021choose} & 0.1401 & 0.0084 & 0.0118 & Transolver \cite{wu2024transolver} & \underline{0.0900} & \underline{0.0057} & \underline{0.0053} \\
geo-FNO \cite{li2023fourier} & 0.1556 & 0.0108 & 0.0138 & \textbf{DiffFluid} & \textbf{0.0497} & \textbf{0.0049} & \textbf{0.0047} \\
\midrule
& & & & \textit{Relative Promotion} & 44.8\% & 14.0\% & 11.3\% \\
\bottomrule
\end{tabular}
}
\vspace{1em}
\caption{Performance comparison based on Navier-Stokes, Darcy, and Airfoil equations, showing relative $L2$ error. Lower values indicate better performance. The best performance is in bold, and the second best is underlined. "/" indicates that the baseline is not applicable to this benchmark.}
\label{tab:combined_performance}
\end{table}
 
\begin{table}[t]
    \centering
    \resizebox{0.9\textwidth}{!}{ 
        \begin{tabular}{p{2.5cm}p{2.5cm}p{2.5cm}p{2.5cm}}  
            \hline
             & $\nu = 1e-3$ & $\nu = 1e-4$ & $\nu = 1e-5$ \\
            Model & $T = 50$ & $T = 30$ & $T = 20$ \\
            & $N = 1000$ & $N = 1000$ & $N = 1000$ \\
            \hline
             \textbf{DiffFluid$\downarrow$} & \textbf{0.0064} & \textbf{0.0372} & \textbf{0.0497} \\
            \hline
            FNO-3D \cite{li2020fourier} & 0.0086 & 0.1918 & 0.1893 \\
            FNO-2D \cite{li2020fourier} & 0.0128 & 0.1559 & 0.1556 \\
            U-Net \cite{li2020fourier} & 0.0245 & 0.2051 & 0.1982 \\
            TF-Net \cite{li2020fourier} & 0.0225 & 0.2253 & 0.2268 \\
            Res-Net \cite{li2020fourier} & 0.0701 & 0.2871 & 0.2753 \\
            \hline
        \end{tabular}
    }
    \vspace{1em}
    \caption{Compared to the series of benchmarks proposed by \cite{li2020fourier}, the performance of the model at different Reynolds numbers. Where $\nu$ represents viscosity, $T$ is the discrete time step, and $N$ is the size of the training dataset.}
    \label{tab:performance_Re}
\end{table}

\subsection{Benchmark and baselines }
Our experiment is based on the Navier-Stokes and Darcy flow equations \cite{li2020fourier}, as well as the airfoil problem using Euler's equations \cite{li2023fourier}. We compare the DiffFluid with several baseline approaches, including neural operators like FNO \cite{li2020fourier}, Transformer-based solvers such as GNOT \cite{hao2023gnot}, and the recent state-of-the-art Transolver \cite{wu2024transolver}.

\subsection{Main results}

As mentioned above, DiffFluid performs exceptionally well in addressing the sharp features within the solution domain of the Navier-Stokes equations, which directly impacts the model's performance. DiffFluid effectively tackles this challenge and is further optimized through multi-resolution noise and multi-loss strategies. Compared to the second-best model, it achieves a performance improvement of 44.8\%. Results can be found in Table \ref{tab:combined_performance}, which presents various performance metrics and comparisons. Additionally, the visualization results of the second-best model Transolver are shown in Figure \ref{fig:ns}.

Additionally, we compare DiffFluid against a series of benchmarks for the Navier-Stokes equations at various Reynolds numbers as proposed by \cite{li2020fourierre}. DiffFluid consistently demonstrates state-of-the-art performance, as highlighted in Table \ref{tab:performance_Re}, underscoring its exceptional capabilities.

To validate the generalization capability of our model, we conduct additional tests on a representative class of stationary fluid dynamical equations known as Darcy flow. The experimental results show that DiffFluid effectively manages smooth solutions like Darcy flow, achieving a state-of-the-art performance improvement of 14.0\%, as detailed in Table \ref{tab:combined_performance}. Moreover, Figure \ref{fig:darcy} displays the visualization results for the second-best model, Transolver.

\begin{table}[ht]
    \centering
    \resizebox{0.9\textwidth}{!}{
        \begin{tabular}{cccccc} 
            \toprule
            Number of Mesh Points & 1,681 & 3,364 & 7,225 & 10,609 & 19,881 \\ 
            (Resolution) & (41$\times$41) & (58$\times$58) & (85$\times$85) & (103$\times$103) & (141$\times$141) \\
            \midrule
            \textbf{DiffFluid} & \textbf{0.0073} & \textbf{0.0052} & \textbf{0.0049} & \textbf{0.0049} & \textbf{0.0054} \\ 
            Transolver \cite{wu2024transolver} & 0.0089 & 0.0058 & 0.0059 & 0.0057 & 0.0062 \\ 
            \midrule
            Relative Error Reduction & 18.0\% & 10.3\% & 16.9\% & 14.0\% & 12.9\% \\ 
            \bottomrule
        \end{tabular}%
    }
    \vspace{1em}
    \caption{A comparison of performance between DiffFluid and Transolver across different mesh resolutions.}
    \label{tab:mesh_points}
\end{table}

We also apply DiffFluid to Darcy Flow tasks at various resolutions and compare it with the second-best model, Transolver \cite{wu2024transolver}, to evaluate the model's adaptability to different resolution fluid dynamical equations. The results indicate that DiffFluid can effectively adapt to tasks across varying resolutions, with additional details provided in Table \ref{tab:mesh_points}. This makes DiffFluid a promising candidate for large-scale industrial applications.

To validate the performance of DiffFluid across different network types, we select a structured mesh for the airfoil problem using Euler's equation, differing from the regular grid used for the Navier-Stokes equations and Darcy flow equations. The experimental results indicate that our model adapts well to this network type and outperforms the second-best by 11.3\% Table \ref{tab:combined_performance}. Furthermore, the visual results of the second-best model, Transolver, can be found in Figure \ref{fig:airfoil}.

\subsection{Ablation study}

As mentioned earlier, the multi-resolution noise and multi-loss strategies effectively help DiffFluid capture sharp features in the fluid dynamical equation solution domain. To this end, we design several ablation experiments for the Navier-Stokes equation task, selecting $\nu = 1e-5$.

\begin{table}[t] 
    \centering
    \resizebox{0.9\textwidth}{!}{ 
        \begin{tabular}{cccc} 
            \toprule
            Gaussian Noise & Annealing Strategy & Multi-resolution Noise & $L_2 \, \text{Error} \downarrow$ \\ 
            \midrule
            $\checkmark$ & $\times$ & $\times$ & 0.0732 \\ 
            $\checkmark$ & $\checkmark$ & $\times$ & 0.0674 \\ 
            $\times$ & $\times$ & $\checkmark$ & 0.0562 \\ 
            $\times$ & $\checkmark$ & $\checkmark$ & \textbf{0.0497} \\ 
            \bottomrule
        \end{tabular}%
    }
    \vspace{1em}
    \caption{A comparison of different noise strategies.}
    \label{tab:noise_strategy} 
\end{table}

\begin{figure}[h]
    \centering
    \resizebox{0.7\textwidth}{!}{\includegraphics{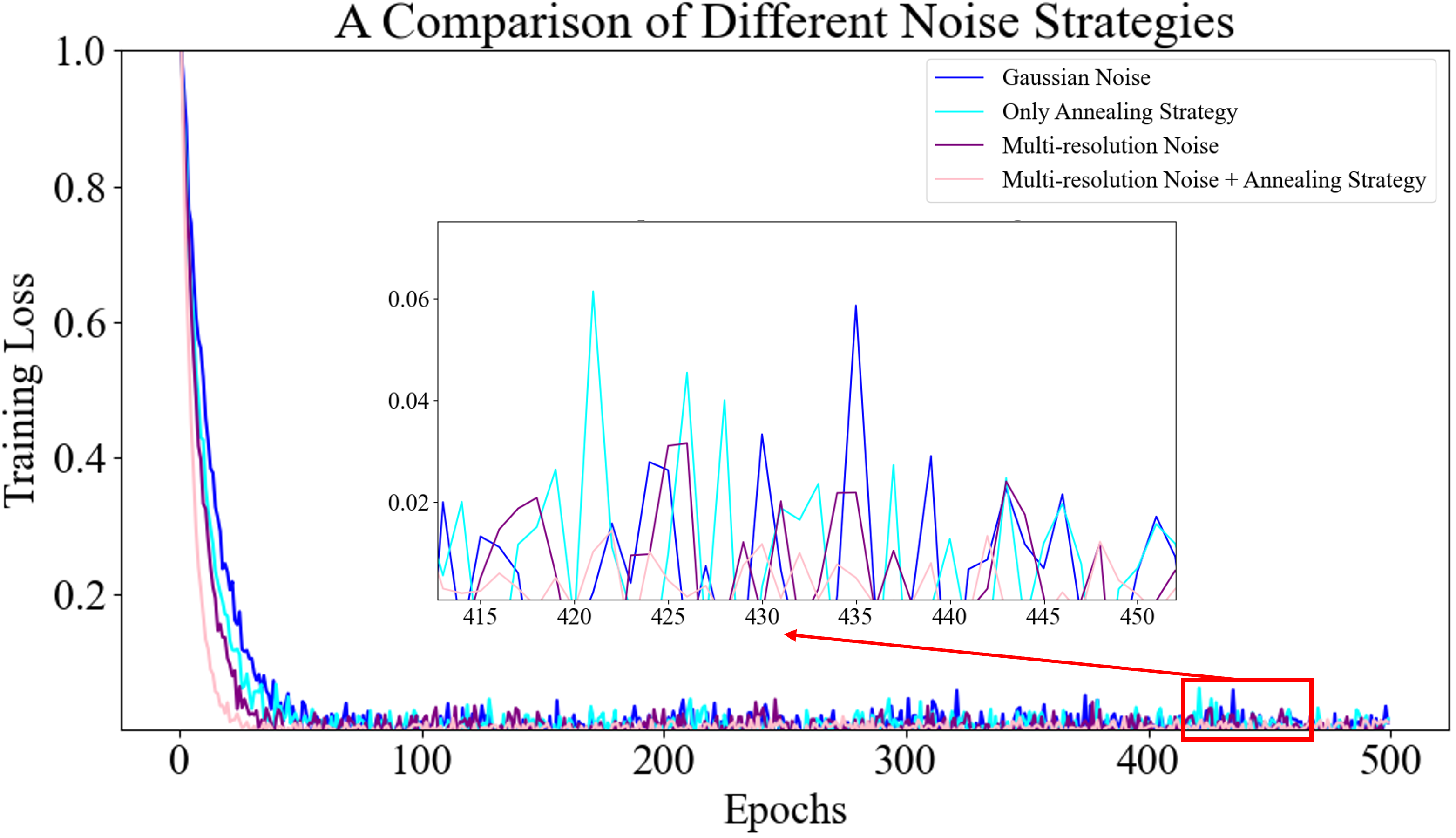}} 
    \caption{A comparison of the effects of different noise strategies on training loss.}
    \label{fig:noise}
\end{figure}

\textbf{Multi-resolution noise} 
We conduct three sets of experiments to verify the impact of noise on the training process and the final solution accuracy. As shown in Figure \ref{fig:noise}, using multi-resolution noise significantly accelerates the fitting speed during training and improves generation accuracy compared to simple Gaussian noise. Additionally, employing an annealing strategy can further enhance both the convergence speed and solution accuracy. Specific results can be found in Table \ref{tab:noise_strategy}.

\textbf{Multi-loss strategy}
To validate the effectiveness of our proposed multi-resolution noise strategy in enhancing the accuracy of solving fluid dynamical equations, we conduct relevant ablation experiments, with the specific results presented in Table \ref{tab:loss_strategy}. The experimental results indicate that using L1 error alone does not yield reliable fluid dynamical equation solutions; however, incorporating L1 error on top of L2 error significantly improves the solution accuracy. Finally, we consider the combined effects of multi-resolution noise and multi-loss functions on the accuracy of DiffFluid solutions. We found that their combined use achieves the best results, as detailed in Table \ref{tab:train_strategy}.

\section{Conclusion and future work}
In this paper,we develop the first Transformer-based diffusion solver for fluid dynamics equations, named DiffFluid. This model effectively captures complex physical and geometric states in fluid dynamics, achieving high-precision solutions.In the validation of representative fluid dynamics problems, such as the Navier-Stokes equations, Darcy's law, and practical applications like wings, DiffFluid achieves consistently state-of-the-art performance. Additionally, we introduced Multi-Resolution Noise and Multi-Loss strategies to further enhance model performance. In the future, we aim to extend DiffFluid to continuous time problems, moving beyond the current slicing limitations to apply it in scenarios resembling continuous videos while improving training speed.

{\small
\nocite{*}
\renewcommand{\bibfont}{\normalfont} 
\renewcommand{\refname}{\vspace{1.5ex}\fontsize{12}{14}\selectfont\bfseries References}
\bibliographystyle{unsrt} 
\bibliography{references} 
}

\clearpage

\appendix

\section{Overview}
In this appendix, we provide detailed content that complements the main paper. Section\ref{sec:implementation-details} elaborates on the implementation details of the experiments, including benchmarks and evaluation metrics. Section \ref{sec:visual-denoising} presents a visual overview of the denoising process. Section \ref{sec:visual-comparison} compares the visualization results of the Darcy flow equation and the airfoil problem with Euler's equation between the previous state-of-the-art Transolver and our DiffFluid. Section \ref{sec:ablation-tables} includes additional tables related to the ablation studies.
\section{Implementation details}
\label{sec:implementation-details}
\subsection{Benchmarks}
We validated the performance of our model on three benchmarks: the Navier-Stokes equations, the Darcy flow equations, and the airfoil problem using Euler's equations. For detailed information about the benchmarks, please refer to Table \ref{tab:selected_benchmark}. Our tests involve the following two types of fluid dynamics equations:

\begin{itemize}
    \item \textbf{Navier-Stokes equations for fluid} \cite{mclean2012understanding}: Navier-Stokes, Airfoil.
    
    \item \textbf{Darcy’s law} \cite{hubbert1956darcy}: Darcy. 
\end{itemize}

The following are the detailed information for each benchmark.

\textbf{Navier-Stokes}\quad This benchmark simulates incompressible viscous flow on a unit torus, where the fluid density is constant and the viscosity is set to $1e-3$, $1e-4$ and $1e-5$. The fluid field is discretized into a $64 \times 64$ regular grid. The task is to predict the future 10 steps of the fluid based on the observations from the previous 10 steps. The model is trained using 1,000 fluid instances with different initial conditions and tested with 200 new samples.

\textbf{Airfoil}\quad This benchmark estimates the Mach number based on airfoil shapes. The input shapes are discretized into a structured grid of $221 \times 51$, and the output is the Mach number at each grid point \cite{li2023fourier}. All shapes are deformations of the NACA-0012 case provided by the National Advisory Committee for Aeronautics. A total of 1,000 different airfoil design samples are used for training, with an additional 200 samples for testing.

\textbf{Darcy}\quad This benchmark is utilized to simulate fluid flow through porous media \cite{li2020fourier}. In the experiment, the process is discretized into a regular grid of $421 \times 421$, and the data is downsampled to a resolution of $85 \times 85$ for the main experiments. The model's input is the structure of the porous medium, while the output is the fluid pressure at each grid point. A total of 1,000 samples are used for training and 200 samples for testing, covering various structures of the medium.

\begin{table}[ht]
    \centering
    \renewcommand{\arraystretch}{2.0} %
    \resizebox{\textwidth}{!}{ 
        \begin{tabular}{p{0.15\textwidth} p{0.18\textwidth} p{0.10\textwidth} p{0.1\textwidth} p{0.18\textwidth} p{0.20\textwidth} p{0.14\textwidth}}
            \toprule
            \Large Geometry & \Large Benchmarks & \Large Dim & \Large Mesh & \Large Input & \Large Output & \Large Dataset \\ \midrule
            \large Regular Grid & \large Navier–Stokes & \large 2D+Time & \large 4,096 & \large Past Velocity & \large Future Velocity & \large (1000, 200) \\ 
            \midrule
            \large Regular Grid & \large Darcy Flow & \large 2D & \large 7,225 & \large Porous Medium & \large Fluid Pressure & \large (1000, 200) \\ 
            \midrule
            \large Structured Mesh & \large Airfoil & \large 2D & \large 11,271 & \large Structure & \large Mach Number & \large (1000, 200) \\ 
            \bottomrule
        \end{tabular}
    }
    \vspace{1em}
    \caption{The benchmarks Navier–Stokes, Darcy Flow, and Airfoil were created by \cite{li2020fourier}. Dim represents the dimension of the dataset, Mesh refers to the size of the discretized grid, and Dataset includes the number of samples in the training and testing sets.}
    \label{tab:selected_benchmark}
    \vspace{1em}
\end{table}

\subsection{Metrics}
To visually demonstrate the state-of-the-art performance of our model and ensure fair comparison with other models, we chose to use relative L2 to measure the error in the physics field. The relative L2 error of the model prediction field $\hat{\phi}$ compared to the given physical field $\phi$ can be calculated as follows:

\begin{equation}
    \text{Relative L2 Loss} = \frac{\|y-\hat{y} \|_2}{\| y \|_2}
\end{equation}

\section{Visualization of denoising process}
\label{sec:visual-denoising}

\begin{figure}[htbp]
    \centering
    \includegraphics[width=1\textwidth]{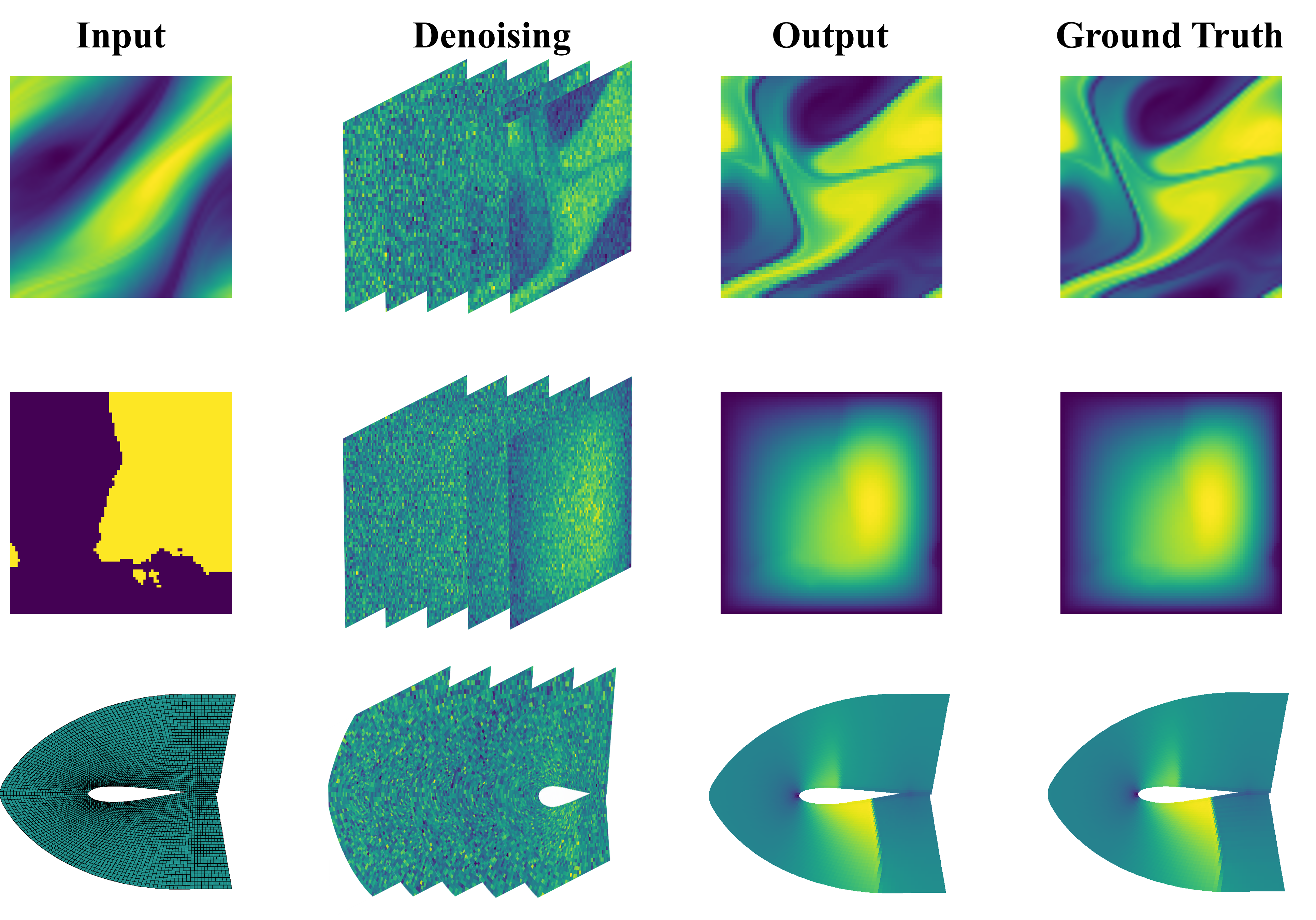} 
    \caption{Denoising performance for fluid dynamics equations: (1) Navier-Stokes Equation; (2) Darcy Flow Equation; (3) Airfoil Problem with Euler’s Equation.}
    \label{fig:intro}
\end{figure}

To provide a clearer visualization of the inputs in the benchmark, the denoising process of DiffFluid, and the comparison between the output results and the actual results, we have visualized this entire process. Please refer to the Figure \ref{fig:intro} for details.

\section{Visual Comparison of Flow Equations}
\label{sec:visual-comparison}

\begin{figure}[htbp]
    \centering
    \includegraphics[width=1\textwidth]{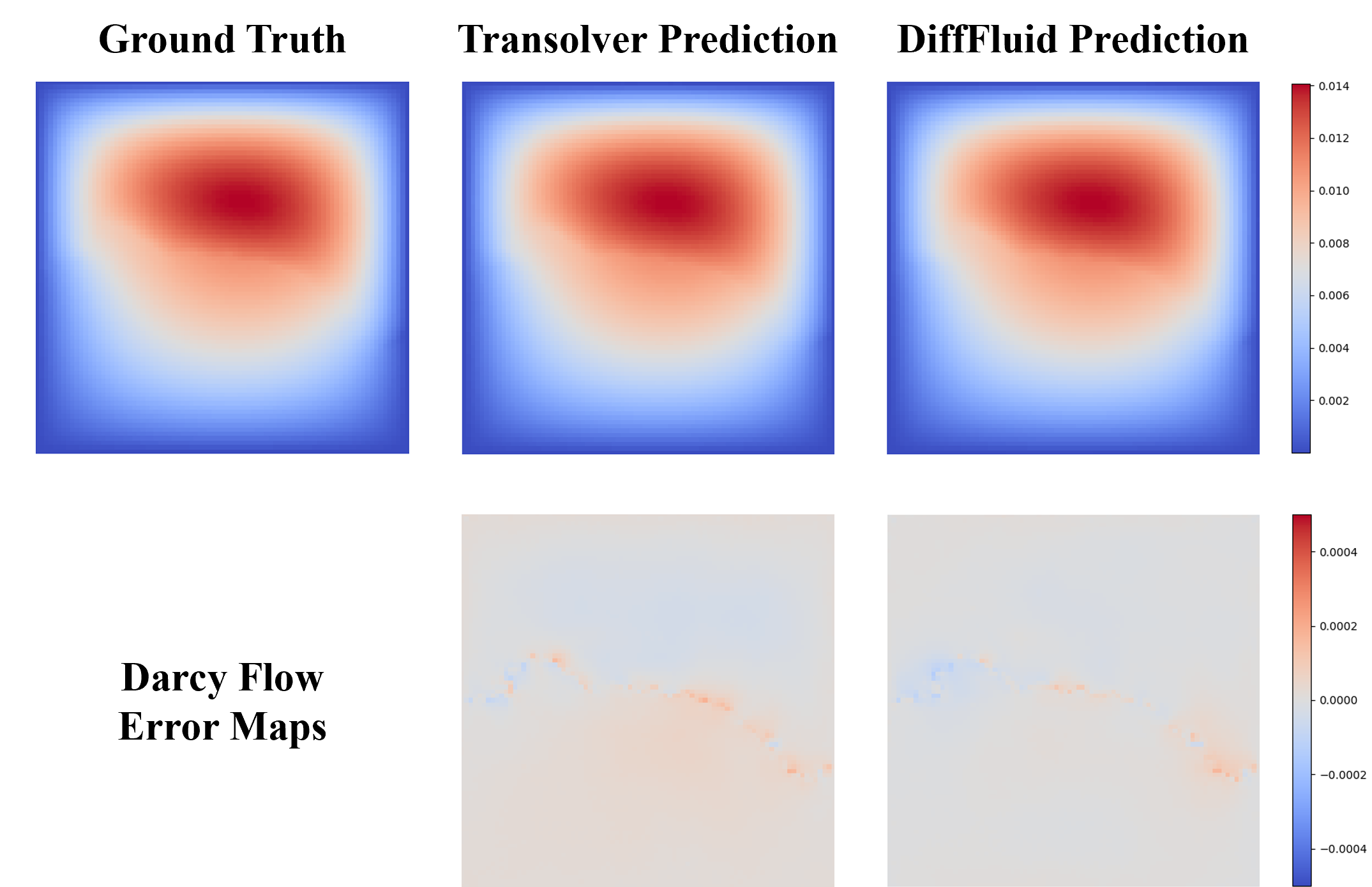} 
    \caption{A comparison of DiffFluid with the previous best model, Transolver, on the Darcy flow equation benchmark. Both prediction results and error maps are provided.}
    \label{fig:darcy}
\end{figure}

\begin{figure}[htbp]
    \centering
    \includegraphics[width=1\textwidth]{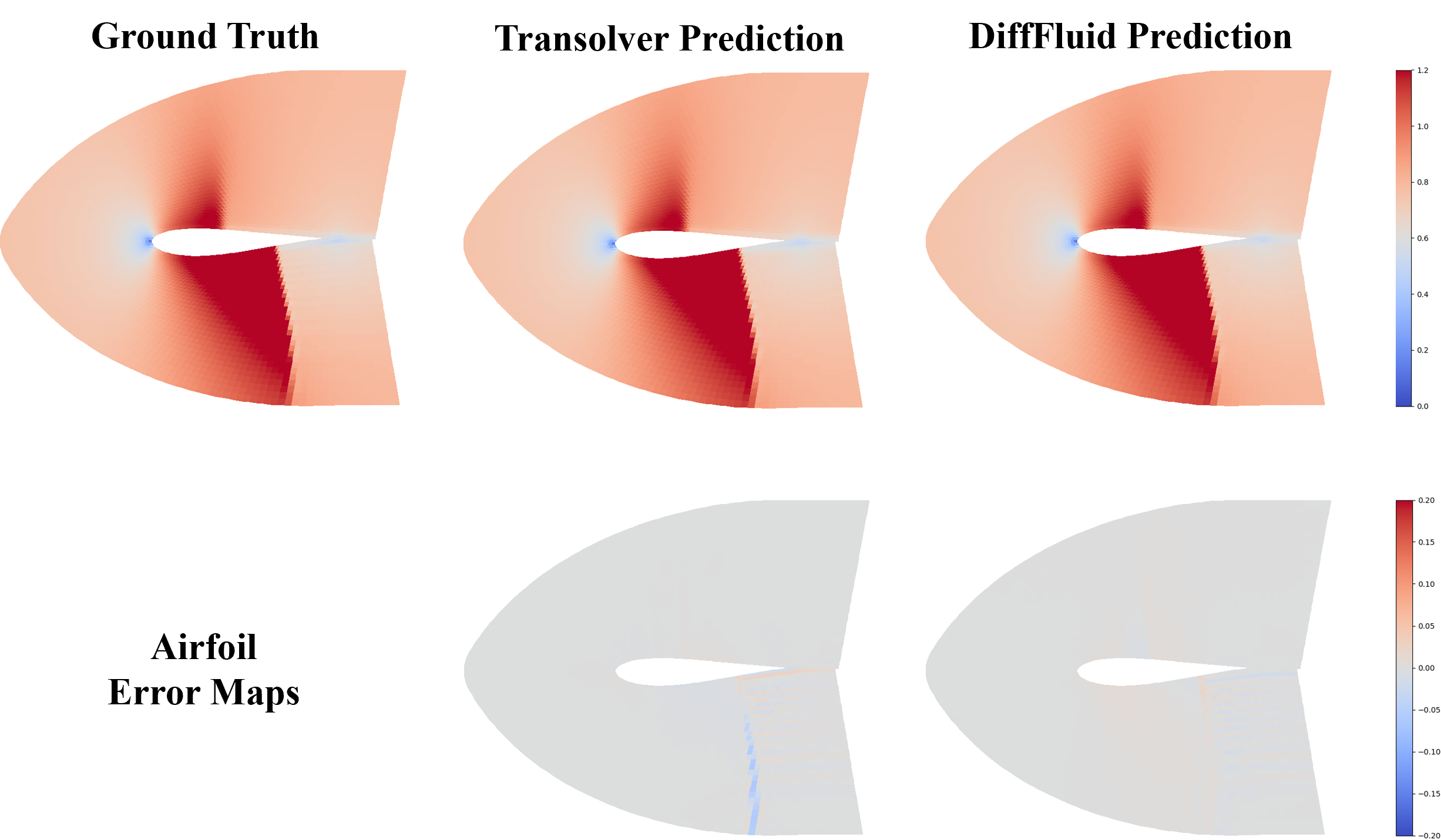} 
    \caption{A comparison of DiffFluid with the previous best model, Transolver, on the Airfoil problem with Euler’s equation benchmark. Both prediction results and error maps are provided.}
    \label{fig:airfoil}
\end{figure}

To clearly and concisely demonstrate the performance improvement of DiffFluid compared to the second-best model, we visualized the results generated by both methods and compared them with the ground truth. In Figures \ref{fig:ns} \ref{fig:darcy} \ref{fig:airfoil}we present comparative graphs for the Navier-Stokes equation, Darcy flow equation, and the airfoil problem based on Euler’s equation.

\clearpage

\clearpage

\section{Supplementary of ablation study}
\label{sec:ablation-tables}

In the ablation study section, we performed several experiments on the Navier-Stokes equation task with a viscosity of $\nu = 1e-5$. We explored different types of noise, various loss function strategies, and combinations of these approaches to assess their impact on performance.

\begin{itemize}
    \item \textbf{Multi-resolution noise}\quad By employing different noise addition strategies, we found that using multi-resolution noise in conjunction with a sampling annealing strategy can significantly enhance model accuracy. For detailed results, please refer to Table \ref{tab:noise_strategy}.
    
    \item \textbf{Multi-loss strategy}\quad By comparing different loss functions used during training, we found that employing a multi-loss strategy can significantly enhance the model's accuracy. For specific results, please refer to Table \ref{tab:loss_strategy}.

    \item \textbf{Comprehensive strategy}\quad Integrating both the multi-resolution noise and multi-loss strategies revealed a synergistic effect that significantly enhances model accuracy. This combination results in improved precision and overall performance. For detailed results, please refer to Table \ref{tab:train_strategy}.
\end{itemize}

\begin{table}[htbp] 
    \centering
    \resizebox{0.7\textwidth}{!}{%
        \begin{tabular}{p{0.25\textwidth} p{0.25\textwidth} c}   
            \toprule
            \quad \quad $L1 \, \text{Loss}$ & \quad \quad $L2 \, \text{Loss}$ & $L2 \, \text{Error} \downarrow$ \\ 
            \midrule
            $\quad \quad \quad \checkmark$ & \quad \quad \quad  $\times$ & 0.3743 \\ 
            $\quad \quad \quad \times$ & \quad \quad \quad $\checkmark$ & 0.0826 \\ 
            $\quad \quad \quad \checkmark$ & \quad \quad \quad $\checkmark$ & \textbf{0.0497} \\  
            \bottomrule
        \end{tabular}%
    }
    \vspace{1em}
    \caption{A comparison of different loss strategies.}
    \vspace{1em}
    \label{tab:loss_strategy}
\end{table}

\begin{table}[htbp] 
    \centering
    \resizebox{0.7\textwidth}{!}{%
        \begin{tabular}{p{0.25\textwidth} p{0.25\textwidth} c}   
            \toprule
            Multi-resolution Noise \quad + Annealing & \quad \quad Multi-loss & $L2 \, \text{Error} \downarrow$ \\ 
            \midrule
            $\quad \quad \quad \quad \times$ & $\quad \quad \quad \quad \times$ & 0.2562 \\
            $\quad \quad \quad \quad \checkmark$ & $\quad \quad \quad \quad \times$ & 0.2273 \\ 
            $\quad \quad \quad \quad \times$ & $\quad \quad \quad \quad \checkmark$ & 0.0532 \\ 
            $\quad \quad \quad \quad \checkmark$ & $\quad \quad \quad \quad \checkmark$ & \textbf{0.0497} \\  
            \bottomrule
        \end{tabular}%
    }
    \vspace{1em}
    \caption{A comparison of different training strategies.}
    \vspace{1em}
    \label{tab:train_strategy}
\end{table}


\end{document}